\documentclass[11pt]{article}

\usepackage[final]{acl}

\usepackage{times}
\usepackage{latexsym}

\usepackage[T1]{fontenc}

\usepackage[utf8]{inputenc}

\usepackage{microtype}

\usepackage{inconsolata}

\usepackage{graphicx}

\usepackage{listings}
\usepackage{xcolor}
\usepackage{todonotes}
\usepackage{booktabs}
\usepackage{multirow}
\usepackage{amssymb}   

\definecolor{codebg}{HTML}{F8F8F8}
\definecolor{kwcolor}{HTML}{0000FF}
\definecolor{strcolor}{HTML}{BA2121}  %
\definecolor{cmtcolor}{HTML}{408080}  %
\definecolor{fncolor}{HTML}{0000FF}   %

\lstdefinestyle{python}{
    language=Python,
    backgroundcolor=\color{codebg},
    basicstyle=\ttfamily\small,
    keywordstyle=\color{kwcolor}\bfseries,
    stringstyle=\color{strcolor},
    commentstyle=\color{cmtcolor}\itshape,
    emph={diversify},          %
    emphstyle=\color{fncolor},
    showstringspaces=false,
    breaklines=false,
    frame=single,
    framesep=4pt,
    rulecolor=\color{gray!30},
}
\lstdefinestyle{json}{
    backgroundcolor=\color{codebg},
    basicstyle=\ttfamily\footnotesize,   %
    string=[s]{"}{"},
    stringstyle=\color{strcolor},
    comment=[l]{:},
    commentstyle=\color{black},
    showstringspaces=false,
    breaklines=true,                     %
    breakatwhitespace=true,              %
    breakindent=2em,                     %
    frame=single,
    framesep=4pt,
    rulecolor=\color{gray!30},
}

\usepackage{xcolor}
\usepackage[most]{tcolorbox}
\usepackage{hyperref}

\usepackage{fontawesome5} %
\usetikzlibrary{arrows.meta} %
\usepackage{enumitem} %

\usepackage{fontawesome5} %
\usetikzlibrary{arrows.meta} %
\usepackage{enumitem} %

\usepackage{tikz}
\newcommand{\barwidth}{2.2cm}
\newcommand{\barheight}{0.4cm}

\definecolor{color1}{HTML}{FF3333}
\definecolor{color2}{HTML}{7EA6E0}
\definecolor{color3}{HTML}{990000}
\definecolor{color4}{HTML}{F5F5F5}

\newcommand{\FullRedBar}[1]{%
\begin{tikzpicture}[baseline=(current bounding box.center)]
  \draw[fill=color1!80] (0,0) rectangle (\barwidth,\barheight);
  \node at (0.5*\barwidth,0.5*\barheight) {\textbf{#1}};
\end{tikzpicture}%
}

\newcommand{\HalfBar}[2]{%
\begin{tikzpicture}[baseline=(current bounding box.center)]
  \draw[fill=color1!80] (0,0) rectangle (0.5*\barwidth,\barheight);
  \draw[fill=color2!80] (0.5*\barwidth,0) rectangle (\barwidth,\barheight);
  \node at (0.25*\barwidth,0.5*\barheight) {\textbf{#1}};
  \node at (0.75*\barwidth,0.5*\barheight) {\textbf{#2}};
\end{tikzpicture}%
}

\newcommand{\ThirdsBar}[3]{%
\begin{tikzpicture}[baseline=(current bounding box.center)]
  \draw[fill=color1!80] (0,0) rectangle (1/3*\barwidth,\barheight);
  \draw[fill=color2!80] (1/3*\barwidth,0) rectangle (2/3*\barwidth,\barheight);
  \draw[fill=color3!65] (2/3*\barwidth,0) rectangle (\barwidth,\barheight);
  \node at (1/6*\barwidth,0.5*\barheight) {\textbf{#1}};
  \node at (0.5*\barwidth,0.5*\barheight) {\textbf{#2}};
  \node at (5/6*\barwidth,0.5*\barheight) {\textbf{#3}};
\end{tikzpicture}%
}

\newcommand{\QuartersBar}[4]{%
\begin{tikzpicture}[baseline=(current bounding box.center)]
  \draw[fill=color1!80] (0,0) rectangle (0.25*\barwidth,\barheight);
  \draw[fill=color2!80] (0.25*\barwidth,0) rectangle (0.5*\barwidth,\barheight);
  \draw[fill=color3!65] (0.5*\barwidth,0) rectangle (0.75*\barwidth,\barheight);
  \draw[fill=color4!80] (0.75*\barwidth,0) rectangle (\barwidth,\barheight);
  \node at (0.125*\barwidth,0.5*\barheight) {\textbf{#1}};
  \node at (0.375*\barwidth,0.5*\barheight) {\textbf{#2}};
  \node at (0.625*\barwidth,0.5*\barheight) {\textbf{#3}};
  \node at (0.875*\barwidth,0.5*\barheight) {\textbf{#4}};
\end{tikzpicture}%
}

\title{\texttt{emb-diversity}: embedding-based diversity measurement}
\title{\texttt{emb-diversity}: \\ A Tool for  Embedding-Based Measurement of Data Diversity}

 \author{Cantao Su, Menan Velayuthan, Esther Ploeger, Dong Nguyen, Anna Wegmann \\
         Department of Information \& Computing Sciences, Utrecht University \\Utrecht, The Netherlands \\
         \small{
   \textbf{Correspondence:} \href{mailto:c.su@uu.nl}{c.su@uu.nl}
 }}

\begin{document}
\maketitle
\begin{abstract}
There is growing evidence that data diversity is crucial for developing fair and robust NLP models.
However, current approaches to measure diversity remain inconsistent and  fragmented:
While there exist a number of tools for measuring the \textit{lexical} diversity of texts, researchers lack standardized tools for quantifying diversity based on \textit{embeddings}.
Embedding-based diversity measures are highly flexible: They work with any embedding model and any data that can be embedded, and are thus applicable to many notions of diversity. %
With \texttt{emb-diversity}, we provide a comprehensive embedding-based diversity measurement tool, spanning a broad range of measures. We demonstrate its potential for several use cases: measuring the stylistic, semantic, language and speaker diversity of datasets.
\noindent\begin{minipage}{\linewidth}
\centering
\href{https://github.com/nlpsoc/emb-diversity/}{\faGithub} 
\href{https://github.com/nlpsoc/emb-diversity/}{/nlpsoc/emb-diversity/}
\end{minipage}
\end{abstract}

\section{Introduction}

A growing body of work suggests that data diversity is important for training NLP models. It can improve  desirable properties such as fairness and robustness as well as data efficiency, and overall benchmark performance  \citep{schiller-etal-2024-diversity,bukharin-etal-2024-data,ling2025diversity,qian-etal-2022-perturbation,xi-etal-2025-samplemix}. 
Additionally, LLM output has also been increasingly studied with respect to diversity. Reasons include the widespread homogenization of generated texts and its consequences \citep{sourati2026homogenizing} and the extensive use of synthetic and augmented data in training models \citep{guo-etal-2024-curious,wang-etal-2025-diversity,schaffelder2026}. Studies have investigated how  output diversity is influenced by factors such as model size, tuning paradigms and training data \citep{guo-etal-2025-benchmarking-linguistic,kirk2024understanding,shypula2025evaluating}, and how output diversity can be improved \citep{chen2026posttraining,deshpande-etal-2025-diverse}.

Despite growing interest in data diversity, many studies do not explicitly measure diversity itself; instead, they assume increased diversity from how datasets are collected or augmented. What's more, even when diversity is measured, approaches remain fragmented and often ad hoc \citep{shaib-etal-2025-standardizing,nguyen-ploeger-2025-need,zhao_icml2024}.
However, understanding diversity's impact and developing methods to improve it ultimately requires fine-grained, standardized approaches to \textit{measuring} diversity  \citep{mitchell2023measuringdata} that researchers can easily adopt. 

An approach which has been gaining popularity is \textbf{embedding-based diversity measurement}, which is highly flexible:
As long as the set of instances 
can be represented in a metric vector space, diversity can be measured %
for the embedded concept \citep{lai-etal-2020-diversity,yu-etal-2022-data}. 
Although embedding models are ubiquitous in NLP and have been trained to represent different relevant concepts, to date, no community-maintained, comprehensive %
tool is available for embedding-based diversity measurement. 
To address this gap, we present \textbf{\texttt{emb-diversity}}, an easy-to-use Python tool for embedding-based diversity measurement.  
In this paper, we demonstrate \texttt{emb-diversity}  for measuring stylistic and semantic (\S\ref{sec:style_casestudy}), language (\S\ref{sec:langdiv_casestudy}) and speaker (\S\ref{sec:speaker_casestudy}) diversity. Although we focus on language data, our tool could also be used for other types of data, as long as they can be represented as vectors, such as graphs \citep{NEURIPS2024_6ac807c9} and images \citep{NEURIPS2023_1f5c5cd0}.

Due to the absence of a single canonical measure for embedding-based diversity, \texttt{emb-diversity} implements a broad range of 22 measures that have been proposed in the literature.
Furthermore, the tool can be easily extended, facilitating the development, use and comparison of future measures.
Consequently, \texttt{emb-diversity} serves both practitioners who wish to measure diversity in their data and researchers seeking to develop, evaluate, or compare embedding-based diversity measures. We release our tool under the MIT license.

\begin{figure*}[h]
    \centering
    \includegraphics[width=\linewidth]{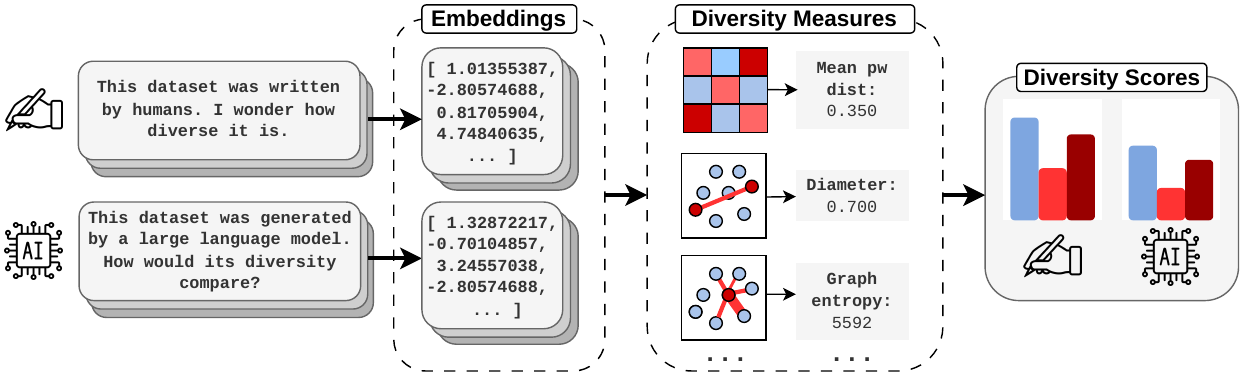}
    \caption{Diagram illustrating the overall \texttt{emb-diversity} pipeline for an example use case of comparing datasets of texts written by humans and by LLMs. Texts are embedded, and a range of diversity scores is calculated for each dataset. The embedding model that is chosen determines what aspect is considered when calculating diversity. %
    }
    \label{fig:overview}
\end{figure*}

\section{Related Work}

Within diversity measurement, lexical diversity has received the greatest attention so far. 
Lexical diversity measures, often based on surface-level patterns such as n-gram overlap, have been implemented by
general data analysis tools  \citep{lohr-hahn-2023-dopa,ramponi-etal-2024-variationist,simig-etal-2022-text} and dedicated software such as TAALED \citep{Kyle15032021} and LexicalRichness \citep{lexicalrichness}. The recently released DELTA tool \citep{esteve-dobrovoljc-2026-delta} focuses on measuring linguistic diversity in
dependency-parsed corpora, but does not focus on embedding-based diversity measurement. %
Most related to \texttt{emb-diversity} is the \texttt{diversity} package \citep{shaib-etal-2025-standardizing}. Although \texttt{diversity} includes 
two embedding-based measures (as of this writing), its main focus differs from ours as it focuses on measuring lexical diversity and repetition in data. 
{Beyond general-purpose tools, many measures are released as standalone reference implementations alongside the papers that introduced them, e.g.\ DCScore \citep{zhu2025measuring}, Hamiltonian diversity \citep{hu2024hamiltonian}, Energy \citep{NEURIPS2024_6ac807c9}, and R\'enyi Kernel Entropy \citep{NEURIPS2023_1f5c5cd0}.
Such code bases each cover a single measure, whereas \texttt{emb-diversity} unifies a broad selection in one tool.}

\paragraph{Concurrent work.} The initial code base for this tool was written as part of \newcite{tao2026}, who collected embedding-based diversity measures, and tested them on a newly introduced benchmark. In this release, we implement them more efficiently to enable more scalable applications, introduce an easy-to-use Python API, add new measures, and aim to maintain the tool over time.

\section{System Overview}

\begin{figure}[t]
\centering
\resizebox{\columnwidth}{!}{%
\begin{tikzpicture}[
  font=\footnotesize,
  box/.style={
    rounded corners=2pt, draw, thick, align=center,
    inner xsep=4pt, inner ysep=2pt, minimum height=6mm, text width=42mm
  },
  side/.style={align=left, font=\scriptsize, text=black!60, text width=26mm},
  flow/.style={-{Stealth[length=2mm]}, thick},
  node distance=6mm,
]
  \node[box, fill=red!12]                    (entry) {Entry points};
  \node[box, fill=cyan!20, below=of entry]   (conv)  {\texttt{measure\_diversity()}};
  \node[box, fill=red!25,  below=of conv]    (meas)  {Measures};
  \node[box, fill=cyan!20, below=of meas]    (embed) {\texttt{resolve\_embeddings()}};
  \node[box, fill=red!35,  below=of embed]   (cache) {Caches};
  \node[box, fill=black!12, below=of cache]  (back)  {Backends};

  \node[side, right=4mm of entry] (entrynote) {CLI \,/\, Python API};
  \node[side, right=4mm of conv]  (convnote)  {convenience function,\\shared by CLI \& API};
  \node[side, right=4mm of meas]  (measnote)  {21 possible measures,\\one function per file};
  \node[side, right=4mm of embed] (embednote) {embed + validate};
  \node[side, right=4mm of cache] (cachenote) {embedding \& pdist,\\checked first};
  \node[side, right=4mm of back]  (backnote)  {ST / HF models,\\run on cache miss};

  \draw[thin, black!50, dotted] (entry.east) -- (entrynote.west);
  \draw[thin, black!50, dotted] (conv.east)  -- (convnote.west);
  \draw[thin, black!50, dotted] (meas.east)  -- (measnote.west);
  \draw[thin, black!50, dotted] (embed.east) -- (embednote.west);
  \draw[thin, black!50, dotted] (cache.east) -- (cachenote.west);
  \draw[thin, black!50, dotted] (back.east)  -- (backnote.west);

  \draw[flow] ([xshift=-18mm]entry.south) -- ([xshift=-18mm]conv.north)
    node[midway,right,font=\scriptsize] {call};
  \draw[flow] ([xshift=-18mm]conv.south)  -- ([xshift=-18mm]meas.north)
    node[midway,right,font=\scriptsize] {dispatch \& lazy loading};
  \draw[flow] ([xshift=-18mm]meas.south)  -- ([xshift=-18mm]embed.north);
  \draw[flow] ([xshift=-18mm]embed.south) -- ([xshift=-18mm]cache.north);
  \draw[flow] ([xshift=-18mm]cache.south) -- ([xshift=-18mm]back.north)
    node[midway,right,font=\scriptsize] {miss};

  \draw[flow] ([xshift=18mm]back.north)  -- ([xshift=18mm]cache.south);
  \draw[flow] ([xshift=18mm]cache.north) -- ([xshift=18mm]embed.south);
  \draw[flow] ([xshift=18mm]embed.north) -- ([xshift=18mm]meas.south);
  \draw[flow] ([xshift=18mm]meas.north)  -- ([xshift=18mm]conv.south);
  \draw[flow] ([xshift=18mm]conv.north)  -- ([xshift=18mm]entry.south);
\end{tikzpicture}%
}
\caption{Architecture of \texttt{emb-diversity}. Input flows top to bottom; every measure funnels through a single embedding-and-validation step, which checks the caches before falling through to the embedding backends. The convenience function can also be skipped in case users want to manipulate parameters of diversity functions directly.}
\label{fig:architecture}
\end{figure}
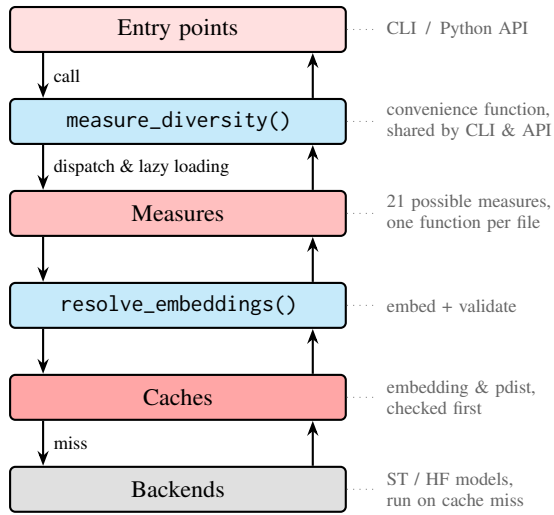

Figure \ref{fig:overview} illustrates the pipeline of a typical use case of \texttt{emb-diversity}: A user may be interested in comparing the diversity of two datasets. When given textual data, \texttt{emb-diversity} first embeds the input. Based on these text embeddings, a range of diversity measures can be calculated, each of which operationalizes the measurement of diversity in a different way. 
Figure \ref{fig:architecture} shows the overall architecture of the tool. In this section, we describe the interfaces through which a user can interact with our tool (\S\ref{sec:interfaces}), the included measures (\S\ref{sec:measures}), and the embedding backend (\S\ref{sec:embedding}). Lastly, we describe optimization steps to improve the speed and scalability of the computations (\S\ref{sec:optimizations}). To keep calculations correct over time, \texttt{emb-diversity} includes {511} tests that run on each pull request merged into the main branch.

\label{sec:architecture}

\subsection{Interfaces}
\label{sec:interfaces}

We anticipate that users of our tool will come from a variety of disciplines and levels of expertise. They may be researchers exploring datasets interactively, engineers integrating the framework into a larger workflow, or practitioners who simply want a numerical assessment for a collection of texts without writing code from scratch. To accommodate these different use cases, we expose our tool through two interfaces: \textbf{(i) a Python API} and \textbf{(ii) a command-line interface (CLI)}.
Below, we briefly discuss these two interfaces. For more details, please refer to our documentation.\footnote{\href{https://nlpsoc.github.io/emb-diversity/}{\nolinkurl{https://nlpsoc.github.io/emb-diversity/}}}

\paragraph{Python library.}
The Python library is the main interface for programmatic use. A measure is called directly on a list of texts or vectors and returns a structured output, making it easy to integrate into notebooks, scripts, and larger pipelines. The same interface accepts either a \emph{diversity axis} (a named concept backed by a default embedding model) or an explicit embedding model, so the user can control \emph{what kind} of diversity is being measured without changing the rest of the call. Figure~\ref{fig:diversify-example-api} shows an example of how the library can be used.

\begin{figure}[t]
\begin{lstlisting}[style=python]
from emb_diversity import (
    measure_diversity)

texts = [
    "I thoroughly enjoy the hair bands.",
    "songs of the 80's are the best.",
    "Hip Hop is going DOWNHILL!!!!!"
]

# Default (semantic) axis.
measure_diversity(texts)

# A single measure on a different axis.
measure_diversity(texts,
    measure="mean_pw_dist",
    diversity_axis="style")

# An explicit embedding model.
measure_diversity(texts,
    embedding_model="all-MiniLM-L6-v2")
\end{lstlisting}
\caption{Example usage of the \textbf{Python API}.}
\label{fig:diversify-example-api}
\end{figure}

\paragraph{Command-line interface.}
The CLI exposes the same functionality without writing code, allowing users to compute diversity directly from a text, CSV, or TSV file. This makes the tool usable for non-programmers as well as easy to integrate into shell scripts and batch jobs. The CLI options mirror the Python API one-to-one (see Figure~\ref{fig:diversify-example-cli}).

\begin{figure}[h]
\begin{lstlisting}[style=python]
# One text per line in texts.txt
emb-diversity measure texts.txt

# A single measure on a different axis
emb-diversity measure texts.txt \
    -m mean_pw_dist --axis style
\end{lstlisting}
\caption{Example usage of the \textbf{CLI}.}
\label{fig:diversify-example-cli}
\end{figure}

\subsection{Diversity Measures}
\label{sec:measures}
We include diversity measures that take a collection of vectors as input (for example, embeddings of a text or audio dataset) and return a float value. %
The tool currently implements 22 measures (Table~\ref{tab:measures}) drawn from the literature, ranging from simple summaries of pairwise distances to measures based on graphs constructed over the embeddings. We build on previous implementations where available. See Appendix \ref{app:implementation-details} for details about the measures' implementations and their references. 
Their formal definitions  and parameters are described in our \href{https://nlpsoc.github.io/emb-diversity/user-guide/measures.html\#measure-descriptions}{documentation}. %
New measures can be added by registering a single function that maps a set of embeddings to a scalar, so the tool can grow as new measures are proposed.
Three measures---Mean Pairwise Distance, Vendi Score, and Graph Entropy---are marked as defaults: the former two are widely used in the literature, whilst Graph Entropy performed well on the benchmark by \citet{tao2026}.

\begin{table}[t]
\centering\small
\setlength{\tabcolsep}{3pt}
\begin{tabular}{@{}l p{4.9cm}@{}}
\toprule
\textbf{Measure} & \textbf{Brief Description} \\
\midrule
\multicolumn{2}{@{}l}{\textit{Distance matrix}} \\
$\bigstar$ Mean PW Dist. & mean of all pairwise distances \\
Sum Pairwise Dist.  & sum of all pairwise distances \\
Energy              & negated mean of inverse distances \\
Span (Medoid)       & mean distance to the medoid \\
KNN                 & mean $k$-th-nearest-neighbour dist. \\
Chamfer Dist.       & mean nearest-neighbour distance \\
Sum Bottleneck      & sum of per-point nearest dists.\ \\
Bottleneck          & smallest pairwise distance \\
Sum Diameter        & sum of per-point farthest dists.\ \\
Diameter            & largest pairwise distance \\
\midrule
\multicolumn{2}{@{}l}{\textit{Distance graph}} \\
MST Dispersion      & total spanning-tree edge length \\
$\bigstar$ Graph Entropy & summed per-node entropy of edge dists. \\
HamDiv              & shortest Hamiltonian circuit length \\
\midrule
\multicolumn{2}{@{}l}{\textit{Kernel matrix}} \\
$\bigstar$ Vendi Score & effective no. of orthogonal directions \\
R\'enyi Kernel Ent. & R\'enyi entropy of kernel spectrum \\
Log-Determinant     & log-volume spanned by embeddings \\
DCScore             & effective no.\ of separable samples \\
\midrule
\multicolumn{2}{@{}l}{\textit{UMAP projection}} \\
Bins Entropy        & bin-occupancy entropy of 2D UMAP \\
Convex Hull (3D)    & convex-hull volume of 3D UMAP proj. \\
\midrule
\multicolumn{2}{@{}l}{\textit{Vector statistics}} \\
Cluster Inertia     & within-cluster sum of squares \\
Span (Centroid)     & spread around the centroid \\
Geo.\ Mean Std.\    & geometric mean of per-dim.\ std.\ \\
\bottomrule
\end{tabular}
\caption{The 22 diversity measures implemented in \texttt{emb-diversity},
grouped by their core computation: Using the pairwise \emph{distance matrix} and graph structures built on it (\emph{distance graph}); using the
\emph{kernel matrix}; using \emph{UMAP projection}s of the embedding space; and
using distance-free embedding \emph{vector statistics}. $\bigstar$ marks the
package defaults.}
\label{tab:measures}
\end{table}

\subsection{Embedding Backend}
\label{sec:embedding}

Encoder models on Hugging Face can be used by providing the matching id to the ``embedding\_model'' parameter to embed a text dataset. Alternatively, users can use a pre-defined diversity axis (currently semantics and style), which automatically uses a default model. See 
our \href{https://nlpsoc.github.io/emb-diversity/user-guide/axes.html}{Github repository} for more information.

\subsection{Computational Optimizations}
\label{sec:optimizations}

Given the many supported measures, all of which rely on neural language models, optimization is a key consideration in our system design. %
To make the tool more computationally efficient, we focused on: (i) enabling the tool to handle long texts efficiently and (ii) 
reducing repeated and shared computation across measures.
Further implementation details can be found in our \href{https://github.com/nlpsoc/emb-diversity/blob/main/optimization_doc.md}{Github repository}.

\paragraph{Handling long texts.}

Most embedding models are constrained by a fixed maximum sequence length, which leads to truncation when input exceeds this limit. %
While using models with longer maximum sequence lengths is a natural solution, this is often not feasible in compute-constrained settings. To address this, we adopt a chunking-and-pooling strategy that enables efficient encoding of long texts. 
Example code of using chunking can be found in the Appendix, Figure~\ref{fig:embediver-example-chunking}.%

\paragraph{Caching.}

Mapping data points to embeddings is a core operation in our tool and can be computationally expensive, especially for large datasets and when running multiple measures over the same inputs. To address this, we implement a caching mechanism to avoid redundant computation across both embedding generation and downstream distance-based computations.
At a high level, we cache (i) text embeddings and (ii) intermediate pairwise distance computations, since both are repeatedly reused across different measures. %

\section{Evaluation}
To evaluate \texttt{emb-diversity}, we test both technical performance and user experience. 
First, we evaluate its ability to surface meaningful differences between controlled datasets (\S\ref{sec:eval_datasets}).
Second, we assess how the implemented measures behave relative to one another in terms of %
runtime (\S\ref{sec:eval_measures}). 
Finally, we report on a small user study about the usability of the tool itself (\S\ref{sec:eval_userstudy}).

\begin{table}[t]
    \centering
    \scalebox{0.95}{
    \begin{tabular}{l|rrr}
    \toprule
         \bf Data Mixture & \bf Vendi & \bf Graph Ent. & \bf MPD \\
     \midrule
         \FullRedBar{N} & 90.10 & 6898.4 & 0.812 \\
         \HalfBar{N}{A} & 137.60 & 6897.5 & 0.917 \\
         \ThirdsBar{N}{A}{G} & 178.77 & 6900.9 & 0.941 \\
         \QuartersBar{N}{A}{G}{Y} & 202.67 & 6902.2 & 0.952 \\
     \bottomrule
    \end{tabular}
    }
    \caption{\texttt{emb-diversity}'s default measures applied to  1,000 lines of %
    different data mixtures from four domains (\colorbox{color1!80}{\bf N}ews, \colorbox{color2!80}{\bf A}rxiv abstracts, \colorbox{color3!65}{\bf G}itHub and \colorbox{color4}{\bf Y}ouTube) from The Common Pile \citep{kandpal2026common}. Higher values indicate higher diversity.}
    \label{tab:dataset_diversity}
\end{table}

\subsection{Comparing Dataset Diversity}
\label{sec:eval_datasets}
To assess the behavior of \texttt{emb-diversity}'s default measures, we apply them to a range of data mixtures with controlled levels of diversity (Table \ref{tab:dataset_diversity}).
We sample 1,000 lines from The Common Pile \citep{kandpal2026common}, starting with one domain (News), incrementally adding four domains (Arxiv abstracts, Github archive, YouTube) while keeping the total data size equal.
We add domains in ascending order of internal diversity (see Appendix \ref{app:div_scores_per_domain}).
To avoid long texts distorting the measures, we only sample lines containing between 201 and 300 characters.
Table \ref{tab:dataset_diversity} shows that measures (using the default, semantic embedding model) behave as expected: the Vendi Score and mean pairwise distance increase monotonically with adding (more diverse) domains. Graph Entropy increases almost monotonically.
In Appendix \ref{app:measure_correlation}, we compare all implemented measures by correlating them on a larger set of Common Pile domains.

\subsection{Comparing Measure Runtime}
\label{sec:eval_measures}

\begin{figure}[t]
    \centering
    \includegraphics[width=\linewidth]{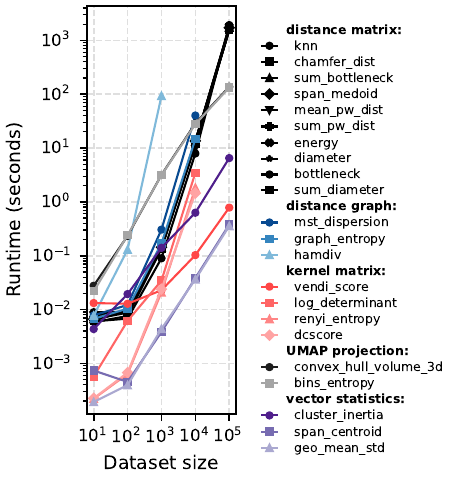}
    \caption{Runtime (seconds) per measure on the same 32-node CPU cluster with 200GB RAM averaged over 5 runs. %
    Runs exceeding 1 hour were aborted. Some measures did not produce results for larger dataset sizes. For example, HamDiv (a traveling salesman algorithm) exceeds one hour at size 10k, while DCscore and MST Dispersion run out of memory at 100k.
    }
    \label{fig:runtime}
\end{figure}

\texttt{emb-diversity} %
enables comparing the runtime of diversity measures on datasets of various sizes. Figure \ref{fig:runtime} shows that some measures, such as the Vendi score, scale well across dataset sizes (synthetic vectors, $d=384$). On the other hand, e.g.,  Hamiltonian diversity is prohibitively expensive to run for larger datasets. This empirical comparison may inform practitioners when choosing a measure.

\subsection{User Study}
\label{sec:eval_userstudy}
Because user‑friendliness is central to the adoption of a practical toolkit, we performed a heuristic evaluation \citep{nielsen1990heuristic} to assess \texttt{emb-diversity}'s usability. Nine NLP researchers examined the tool's first release across both interfaces, including the Python API (both locally and through Google Colab) and the command‑line interface.
They were asked to conduct six pre-determined tasks, such as installing the tool, calculating MST dispersion over a small text dataset, and comparing the diversity of two datasets.
While performing these tasks, they were asked to evaluate \texttt{emb-diversity} according to six usability heuristics \citep{nielsen1990heuristic}. These spanned system status visibility, consistency, error prevention, flexibility/efficiency of use, error recovery, documentation. Further details are in Appendix \ref{app:user_study}.

The user study retrieved 71 comments in total. While the feedback was mostly positive, it also revealed areas for improvement, especially with regard to the documentation.
For example, one participant noted that \textit{
``The documentation does not provide much guidance on how to interpret diversity scores when comparing multiple datasets or determine whether a difference is meaningful.''}.
We added a section in the documentation about \href{https://nlpsoc.github.io/emb-diversity/user-guide/measures.html#interpreting-the-scores}{interpreting the scores}. Another suggestion that we incorporated, was to include an \href{https://nlpsoc.github.io/emb-diversity/user-guide/measures.html#measures}{explicit mapping} between between standard measures names (e.g. MST Dispersion) and their corresponding function (e.g. mst\_dispersion).

\begin{figure*}[h]
    \centering
    \begin{minipage}{0.4\linewidth}
        \centering
        \includegraphics[width=\linewidth]{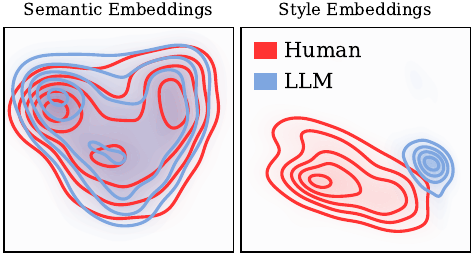}
    \end{minipage}%
    \begin{minipage}{0.55\linewidth}
        \centering\small
        \begin{tabular}{l rr | rr}
        \toprule
            & \multicolumn{2}{c}{\bf Semantic} & \multicolumn{2}{c}{\bf Style} \\
            \cmidrule(lr){2-3} \cmidrule(lr){4-5}
            \bf Measure & \bf Human & \bf LLM & \bf Human & \bf LLM \\
        \midrule
            Vendi          &  89.0%
                            &  83.8%
                            & 3.3%
                           & 1.57 \\
            Graph Entropy  & 5539.1  & 5538.9  & 5369.2 & 5319.6 \\
            Mean PW Dist   &  0.825  &  0.824  & 0.348  & 0.089 \\
        \bottomrule
        \end{tabular}
    \end{minipage}
    \caption{\textbf{Left:} Contours of the density plot of generated and human-written essays on the same topics with a semantic and a style embedding model. \textbf{Right:} Embedding-based diversity measures  over the same dataset. %
    Note that the embedding models for semantics and style are different and scores are thus not comparable across axes.}
    \label{fig:style-diversity}
\end{figure*}

\section{Case Studies}
\label{sec:case_studies}

\noindent
Embedding-based diversity measures have most often been applied to text, using semantic embeddings to capture semantic diversity \citep{tevet-berant-2021-evaluating, han-etal-2022-measuring}. However, embedding-based diversity measurement is a general technique, applicable to any data modality (e.g., text or audio data) and any diversity axis (e.g., semantic or style diversity) for which an embedding model is available.
In this section, we apply diversity measures to text data, typological features, and audio data to measure stylistic, semantic, language and speaker diversity.
{We release the scripts for these examples under the ``examples'' folder in our tool.%
}

\subsection{%
Stylistic vs. Semantic Diversity}
\label{sec:style_casestudy}

Investigating the diversity of LLM-generated texts has been of increasing interest recently \cite{guo-etal-2024-curious,schaffelder2026,chen2026posttraining}. In this case study, we use \texttt{emb-diversity} to compare the stylistic and semantic diversity between texts written by LLMs and texts written by humans on the same essay prompts. %
We sample $\approx800$ essays for both humans and LLMs on the same topics from the %
GEDE corpus \cite{gehring2025gede}, for details see Appendix \ref{app:gede}. %
We use the default style and semantic embeddings from our tool: \href{https://huggingface.co/AnnaWegmann/Style-Embedding}{\textsc{AnnaWegmann/Style-Embedding}} %
\cite{wegmann-etal-2022-author} and \href{https://huggingface.co/sentence-transformers/all-mpnet-base-v2}{\textsc{all-mpnet-base-v2}} %
\cite{reimers-gurevych-2019-sentence}. The resulting diversity values are displayed in Figure~\ref{fig:style-diversity}. While the semantic embeddings show similar diversity values between the two corpora, style is clearly more diverse in the human written essays. %

\subsection{%
Language Diversity}
\label{sec:langdiv_casestudy}

Measuring the diversity of a language sample has long been of interest, both within multilingual NLP (e.g. \citealp{moran-2016-acqdiv,samardzic-etal-2024-measure,ploeger-etal-2026-principled}) and linguistics (e.g. \citealp{dahl2008exercise,stoll2013capturing,miestamo2016sampling}).
In this case study, we calculate language diversity using language vectors provided by the \texttt{lang2vec} toolkit \citep{littell2017uriel}. Specifically, we use \texttt{syntax\_knn} vectors, which reflect structural properties of languages, with missing values imputed from similar languages.
In Table \ref{tab:lang-emb-scores} we contrast two language selections: one that intuitively contains structurally  similar languages (Dutch, Danish, Swedish, German; all Indo-European, Germanic languages), and one that is intuitively more linguistically diverse (Dutch (Indo-European), Hungarian (Uralic), Basque (Basque) and Yoruba (Niger-Congo)).\footnote{Family information from \url{https://wals.info}.}
The results show that the latter sample retrieves higher diversity scores than the former, across all three default measures.

\begin{table}[h]
    \centering
    \small
    \begin{tabular}{lrr}
    \toprule
          & \bf \texttt{\{nld, dan,}   & \bf \texttt{\{nld, hun,}  \\
        \bf Measure & \bf \texttt{\{swe, deu\}} & \bf \texttt{eus, yor\}} \\
     \midrule
        Vendi & 1.3 & 2.4 \\
        Graph Entropy & 4.0 & 4.3 \\
        Mean PW Dist & 0.079 & 0.375 \\
     \bottomrule
    \end{tabular}
    \caption{Scores of three embedding-based diversity measures, calculated over an intuitively
    low-diversity selection (nld, dan, swe, deu) and an intuitively 
    high-diversity language selection (nld, hun, eus, yor).}
    \label{tab:lang-emb-scores}
\end{table}

\subsection{Speaker Diversity with Audio Data}
\label{sec:speaker_casestudy}

Our tool operates on any collection of vectors, regardless of how they are produced. To demonstrate this further, we apply it to audio data: We embed audio utterances with a speaker verification model \href{https://huggingface.co/speechbrain/spkrec-ecapa-voxceleb}{\texttt{speechbrain/spkrec-ecapa-voxceleb}} %
\citep{desplanques2020ecapa}, %
which maps a recording to a fixed-dimensional vector encoding the speaker's voice identity (e.g., pitch).
We use the CMU ARCTIC corpus\footnote{\url{http://www.festvox.org/cmu_arctic/}}%
\citep{kominek2004cmu}, in which every speaker reads the same set of sentences. From it, we construct four datasets with $k=2,3,4,5$ speakers respectively. Each dataset contains the full 327 sentence set exactly once, with sentences assigned to the $k$ speakers in round-robin order: speaker one reads the first sentence, speaker two the second, and so on. %
As shown in Table \ref{tab:speaker-diversity}, the measured diversity increases with the number of speakers. %

\begin{table}[h]
    \centering\small
    \begin{tabular}{l rrrr}
    \toprule
        & \multicolumn{4}{c}{\bf Number of speakers} \\
        \cmidrule(lr){2-5}
        \bf Measure & \bf 2 & \bf 3 & \bf 4 & \bf 5 \\
    \midrule
        Vendi          & 12.6 & 20.7 & 24.8 & 26.5 \\
        Graph Entropy  & 1847.9 & 1865.1 & 1872.0 & 1875.9 \\
        Mean PW Dist            & 0.640 & 0.774 & 0.805 & 0.817 \\
    \bottomrule
    \end{tabular}
    \caption{Embedding-based diversity of a fixed set of CMU ARCTIC sentences as the number of distinct speakers increases from two to five. %
    }
    \label{tab:speaker-diversity}
\end{table}

\section{Conclusion}

We present \texttt{emb-diversity}, an easy-to-use Python tool for embedding-based diversity measurement covering 22 different diversity measures.  %
We hope that our Python tool provides an accessible foundation to investigate data diversity from different perspectives across research fields and applications.

\paragraph{Ongoing Developments.} We are actively improving and maintaining the \texttt{emb-diversity} Python tool. We will include new embedding-based diversity measures as they are developed, and will further work on optimizing the calculation of measures.

\newpage 
\bibliography{custom}

\clearpage
\appendix

\section{Implemented Measures} \label{app:implementation-details}

Our implementations fall into three categories. First, for the Vendi Score \citep{friedman2022vendi}, including its order-$q$ generalization \citep{pasarkar2024cousins}, we call the official \texttt{vendi-score} package directly. Second, five measures are independent reimplementations, cross-checked against the code released with the respective papers: Energy \citep{NEURIPS2024_6ac807c9}, Log-Determinant \citep{kulesza2012determinantal, wang2024diversity}, DCScore \citep{zhu2025measuring}, R\'enyi Kernel Entropy \citep{NEURIPS2023_1f5c5cd0}, and HamDiv \citep{hu2024hamiltonian}. Third, the remaining measures are implemented from their formal definitions: mean and sum pairwise distance \citep{tevet-berant-2021-evaluating, zhang2024improving, lee2023beyond, cox2021directed, yu-etal-2022-data, yang-etal-2025-measuring}, Chamfer distance \citep{cox2021directed, zhang2025evaluating}, KNN distance \citep{yang-etal-2025-measuring}, convex hull volume \citep{yu-etal-2022-data}, span around the centroid and medoid \citep{cox2021directed}, geometric mean of per-dimension standard deviations \citep{lai-etal-2020-diversity}, diameter, bottleneck and their sum \citep{mironov2025measuring, xie2023much}, , cluster inertia \citep{yang-etal-2025-measuring, du2019boosting}, graph entropy \citep{yu-etal-2022-data}, bins entropy \citep{cox2021directed, yang-etal-2025-measuring}, and MST dispersion \citep{atwal2025privacy}.

\section{Optimization}

Further details about how the tool handles long texts and our caching mechanism 
can be found at \url{https://nlpsoc.github.io/emb-diversity/notes.html}.
An example of how chunking can be used with \texttt{emb-diversity} is shown in Figure~\ref{fig:embediver-example-chunking}.

\begin{figure}[h]
\begin{lstlisting}[style=python]
from emb_diversity import (
    measure_diversity, mean_pw_dist)

# Long documents that exceed the
# model's maximum sequence length.
texts = [doc1, doc2, doc3]

# From the main function:
measure_diversity(texts,
    measure="mean_pw_dist",
    diversity_axis="semantic",
    chunking_kwargs={"chunking": True,
                     "chunks": 5,
                     "pooling": "mean"})

# Or directly from an individual measure:
mean_pw_dist(texts,
    diversity_axis="semantic",
    chunking_kwargs={"chunking": True,
                     "chunks": 5,
                     "pooling": "mean"})
\end{lstlisting} \vspace{-.5\baselineskip}
\caption{Chunking is available both from the main \texttt{measure\_diversity}
function and from individual measures, via the same \texttt{chunking\_kwargs}
argument.}
\label{fig:embediver-example-chunking}
\end{figure}

\section{Stylistic Diversity} \label{app:gede}

    We take the ``Task'' and ``Human'' contribution levels from the GEDE corpus  \cite{gehring2025gede}, which correspond to fully human written texts and fully LLM written texts. We match the texts based on the provided ``question\_id''. We drop essay topics that have not been performed by both humans and LLMs. Some tasks have been performed by more than one LLM or human, in which case we sample one essay each. This leaves us with exactly 826 essays each, so a total of 1652 essays.\footnote{To reproduce: \url{https://github.com/nlpsoc/emb-diversity/blob/main/examples/gede_diversity.py}}

\begin{figure*}[h]
    \centering
    \includegraphics[width=.9\linewidth]{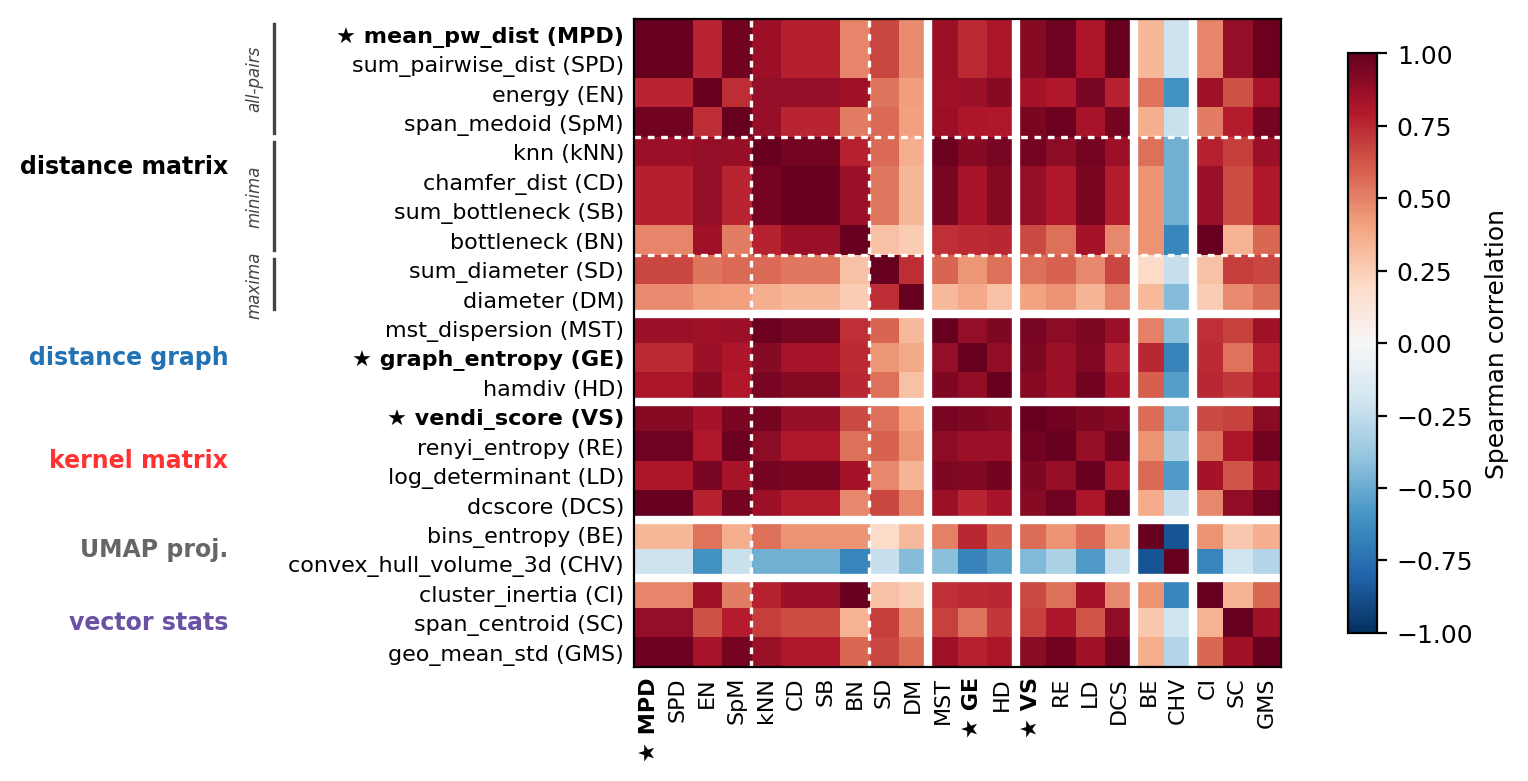}%
    \caption{Pairwise correlations across measures. %
    The correlation scores were calculated on samples from 17 Common Pile domains. Measures are grouped by how they were calculated. Find the script to plot these correlations on our \href{https://github.com/nlpsoc/emb-diversity/blob/main/assets/correlation_grouped.py}{\nolinkurl{GitHub}}.
    }
    \label{fig:corr_heatmap}
\end{figure*}

\section{Details of Heuristic Evaluation}
\label{app:user_study}

Participants were asked to conduct the following tasks:
\begin{enumerate}[itemsep=0pt, parsep=0pt, topsep=0pt]
    \item Install tool
    \item Calculate the diversity over a dataset which consists of three vectors
    \item Calculate the MST dispersion over a small text dataset
    \item Calculate the stylistic diversity over a text dataset
    \item Calculate the diversity over a dataset from HuggingFace
    \item Compare the diversity of two datasets
\end{enumerate}
\vspace{0.5em}

\noindent
We asked participants to give feedback according to the following usability heuristics, defined by \citet{nielsen1990heuristic}. We added short explanations of each heuristic, which were adapted from \url{https://www.nngroup.com/articles/how-to-conduct-a-heuristic-evaluation/}. We selected only heuristics applicable to our toolkit (i.e., excluding concepts that are mostly relevant to GUIs). Due to time constraints, not every participant finished every task.
\vspace{0.5em}
\begin{enumerate}[itemsep=0pt, parsep=0pt, topsep=0pt]
    \item Visibility of System Status (\textit{"The design should always keep users informed about what is going on, through appropriate feedback within a reasonable amount of time."})
    \item Consistency and Standards (\textit{"Users should not have to wonder whether different words, situations, or actions mean the same thing. Follow platform and industry conventions."})
    \item Error Prevention (\textit{"Good error messages are important, but the best designs carefully prevent problems from occurring in the first place. Either eliminate error-prone conditions, or check for them and present users with a confirmation option before they commit to the action."})
    \item Flexibility and Efficiency of Use (\textit{"Shortcuts — hidden from novice users — may speed up the interaction for the expert user such that the design can cater to both inexperienced and experienced users. Allow users to tailor frequent actions."})
    \item Help Users Recognize, Diagnose, and Recover from Errors (\textit{"Error messages should be expressed in plain language (no error codes), precisely indicate the problem, and constructively suggest a solution."})
    \item Help and Documentation (\textit{"It’s best if the system doesn’t need any additional explanation. However, it may be necessary to provide documentation to help users understand how to complete their tasks."})
\end{enumerate}

\section{Diversity Scores per Common Pile Domain}
\label{app:div_scores_per_domain}

Table \ref{tab:div_scores_per_domain} displays the diversity scores for the domains in Table 
\ref{tab:dataset_diversity} sourced from \url{https://huggingface.co/common-pile}.

\begin{table}[h]
    \centering
    \scalebox{0.8}{
    \begin{tabular}{l|ccc}
    \toprule
         \bf Domain ID & \bf Vendi & \bf Graph Ent. & \bf MPD \\
     \midrule
         arxiv\_abstracts\_filtered & 106.0 & 6897.6 &0.854\\
         youtube\_filtered & 192.9 & 6902.6 &0.936\\
         news\_filtered & 90.1 & 6898.4 & 0.812 \\
    github\_archive\_filtered & 141.3 & 6900.4 & 0.883\\
     \bottomrule
    \end{tabular}
    }
    \caption{\texttt{emb-diversity}'s default measures applied to  1,000 lines (with length 201-300 characters) sampled from four domains from Common Pile \citep{kandpal2026common}.}
    \label{tab:div_scores_per_domain}
\end{table}

\section{Correlation among Measures}
\label{app:measure_correlation}
We sample 100 lines from all Common Pile domains from which 100 lines of length 201-300 characters could be extracted within five minutes (excluding long-text domains). %
This results in 17 samples, from
17 \texttt{\_filtered} Common Pile domains: {arxiv\_abstracts}, {youtube}, {news}, {regulations}, {foodista}, {github\_archive},
{wikiteam},
{wikimedia},
{usgpo},
{ubuntu\_irc},
{stackv2\_edu},
{stackexchange},
{pubmed},
{data\_provenance\_initiative},
{cccc},
{biodiversity\_heritage\_library},
{stackv2\_html}. We apply each diversity measure to all samples, and compute pairwise correlations.
Figure \ref{fig:corr_heatmap} shows the results: There are strong correlations between many diversity measures, while correlations with certain measures (e.g. Convex Hull Volume, Diameter) are weaker. Note that we do not expect uniformly high correlations. There exist many different measures because they are designed to measure different aspects of diversity.

\end{document}